\documentclass{article}

     \usepackage[nonatbib, preprint]{neurips_2020}

\usepackage[utf8]{inputenc} 
\usepackage[T1]{fontenc}    
\usepackage{hyperref}       
\usepackage{url}            
\usepackage{booktabs}       
\usepackage{amsfonts}       
\usepackage{nicefrac}       
\usepackage{microtype}      

\usepackage{amsmath}
\usepackage{xfrac}

\usepackage{algorithm} 
\usepackage{algpseudocode} 

\title{Learning Boost by Exploiting the Auxiliary Task in Multi-task Domain}

\author{%
  Jonghwa Yim \\
  Visual Solution Lab.\\
  Samsung Electronics\\
  Suwon-si, Samsung-ro 129 \\
  \texttt{jonghwa.yim@samsung.com} \\
  \And
  Sang Hwan Kim\thanks{This work was mainly developed when the author was an intern at Samsung Electronics.} \\
  School of Electrical Engineering\\
  KAIST \\
  Daejeon, Daehak-ro 291 \\
  \texttt{kshwan0227@kaist.ac.kr} \\
}

\begin{document}

\maketitle

\begin{abstract}
  Learning two tasks in a single shared function has some benefits. Firstly by acquiring information from the second task, the shared function leverages useful information that could have been neglected or underestimated in the first task. Secondly, it helps to generalize the function that can be learned using generally applicable information for both tasks. To fully enjoy these benefits, Multi-task Learning (MTL) has long been researched in various domains such as computer vision, language understanding, and speech synthesis. 
  While MTL benefits from the positive transfer of information from multiple tasks, in a real environment, tasks inevitably have a conflict between them during the learning phase, called negative transfer. The negative transfer hampers function from achieving the optimality and degrades the performance. To solve the problem of the task conflict, previous works only suggested partial solutions that are not fundamental, but ad-hoc. A common approach is using a weighted sum of losses. The weights are adjusted to induce positive transfer. Paradoxically, this kind of solution acknowledges the problem of negative transfer and cannot remove it unless the weight of the task is set to zero. Therefore, these previous methods had limited success. In this paper, we introduce a novel approach that can drive positive transfer and suppress negative transfer by leveraging class-wise weights in the learning process. The weights act as an arbitrator of the fundamental unit of information to determine its positive or negative status to the main task.
\end{abstract}

\section{Introduction}

Multi-task Learning (MTL) is believed to enhance the performance of more than one task by increasing the total amount of information and sharing common and interchangeable information via a shared function, which is called positive transfer. Unlike lofty ideals of MTL, however, from a view of optimization, different tasks regulate each other, causing task conflict, and therefore, preventing them from achieving an optimal point with the best use of all the information. Therefore, previous works in the MTL domain majorly focused on how to negotiate between tasks while training the shared function. One of the conventional methods is using a weighted linear sum of task losses to solve the task conflict. Previous works~\cite{kendall2018multi, chen2018gradnorm, sener2018multi, lin2019pareto} focused on how to balance such weights in the training process. While the weighted sum of task losses admits the presence of negative transfer, such a method could not entirely subdue the negative transfer because some information required by one task can be malicious to the other. If there is an approach suppresses the negative information but accelerates the beneficial one for the main task that we want to enhance, the main task will get to the Pareto optimal point where it fully exploits all the available information only for itself. 

Recent works in the MTL domain well approximate the Pareto optimality so that the shared function is appropriate for performing two or more given tasks. Even when there is one main task that we want to promote, the common belief is that MTL outdoes single-task learning. It can be generally true, but unfortunately, some task pairs are never getting much better than single-task learning, due to the phenomenon known as a negative transfer. Moreover, previous works rarely focused on one main task but tried to optimize all given tasks. More recently, few studies suggest methods to put preference on one task over others. One such recent study, Pareto MTL~\cite{lin2019pareto}, supports the bias vector that adjusts trade-offs among tasks. However, the performance is limited due to the weighted linear sum of losses. Therefore, to meet the industrial needs, there needs another solution to support choosing the one main task that we want to focus on through MTL.

The general belief in MTL is that it can boost the performance through the information gain and obtaining generality by obtaining the feature that works on many tasks. However, without dealing with the task conflict, we cannot achieve the optimal performance for each task. In this paper, we delve into the unit of information, whether it is positive or negative for the task on which we focus. Then we suggest the class-wise weight to deal with the unit of information.

\section{Related Work}

\paragraph{Multi-task Learning (MTL)}
The setting of MTL consists of multiple learning tasks with the hope that the overall performance would be improved by leveraging the knowledge of some or all tasks. MTL has been studied before the advent of deep learning~\cite{caruana1997multitask, bakker2003task}. However, excellent single-task performance and outstanding applicability of deep learning sparked a huge interest in MTL as it can be a good solution for insufficient data problem and fully utilize useful information contained in multiple tasks. Today, MTL has many applications in various fields, including computer vision~\cite{abdulnabi2015multi, li2014heterogeneous, bischke2019multi, ranjan2019competitive}, natural language processing~\cite{collobert2008unified, luong2015multi, subramanian2018learning}, and bioinformatics~\cite{liu2010multi, he2016novel, wang2019cross}.

\paragraph{Transfer Learning}
Transfer learning is a well-known machine learning method in which a trained model is reused on another related task as a starting point. It has been widely applied in various machine learning problems as it not only accelerates the training of a model but also generalizes a model specialized on one problem to other relevant problems. A recent paper~\cite{zamir2018taskonomy} suggested the combination of tasks to boost the learning performance by considering task transferability between tasks and maximizing the advantage of transfer learning. Our research is at the intersection of MTL and Transfer learning. Although we basically follow the setting of previous MTL papers, our model can be regarded as a transfer learning in terms of the relationship between main and auxiliary tasks. In this paper, we focus on the unit of information from the auxiliary tasks to investigate its usefulness to perform the main task learning in the MTL domain. Then we validate our idea on the most popular visual problems: semantic segmentation and depth regression.

Among various approaches to MTL problems, the linear sum of task losses is one of the most common methods to represent the objective function of MTL. With the summed loss, the deep learning algorithm can update parameters of the shared function and task function toward reducing overall task losses. Uniform or manually tuned loss weights was adopted in early studies~\cite{sermanet2013overfeat, kokkinos2017ubernet, eigen2015predicting} to learn multiple tasks simultaneously. Later, it was confirmed that the accuracy of the MTL model highly depends on the proper choice of loss weights by Kendall et al.~\cite{kendall2018multi}. In order to avoid exhaustive and expensive grid search for optimal loss weights, Chen~\cite{chen2018gradnorm} proposed an adaptive algorithm that could optimize task weights by balancing training speeds of multiple tasks. Moreover, Kendall~\cite{kendall2018multi} developed a principled multi-task loss function by applying homoscedastic uncertainty as task-dependent weights. Although these papers~\cite{chen2018gradnorm, kendall2018multi} present an efficient way to reach optimal task weights, its solution is valid only if tasks are correlated.

A more recent study~\cite{sener2018multi} searches for Pareto optimal points by formulating multi-task learning as multi-objective optimization (MOO-MTL), which makes MTL problems even with competing tasks solvable. This idea is more generalized later by Lin et al.~\cite{lin2019pareto} in a way that a set of multiple Pareto solutions with appropriate trade-offs can be obtained to satisfy MTL practitioners’ needs. MOO-MTL~\cite{sener2018multi} and Pareto MTL~\cite{lin2019pareto} suggested a new paradigm to get optimal task weights with proper trade-off among heterogeneous tasks adhering to the linear combination of objective losses. However, weighted sum loss even at Pareto optimal points has some limits due to its assumption: the contribution of each task to total loss is computed without internal weight neglecting internal effect within a task, which hinders a model from taking full advantage of detailed information contained in various tasks. In this paper, therefore, we propose a novel algorithm that can fully maximize positive transfer and minimize negative transfer by optimizing internal class-wise weights.

\section{Task Conflict Analysis in Multi-task Domain}

\begin{figure}[ht]
  \centering
  \includegraphics[width=1\linewidth]{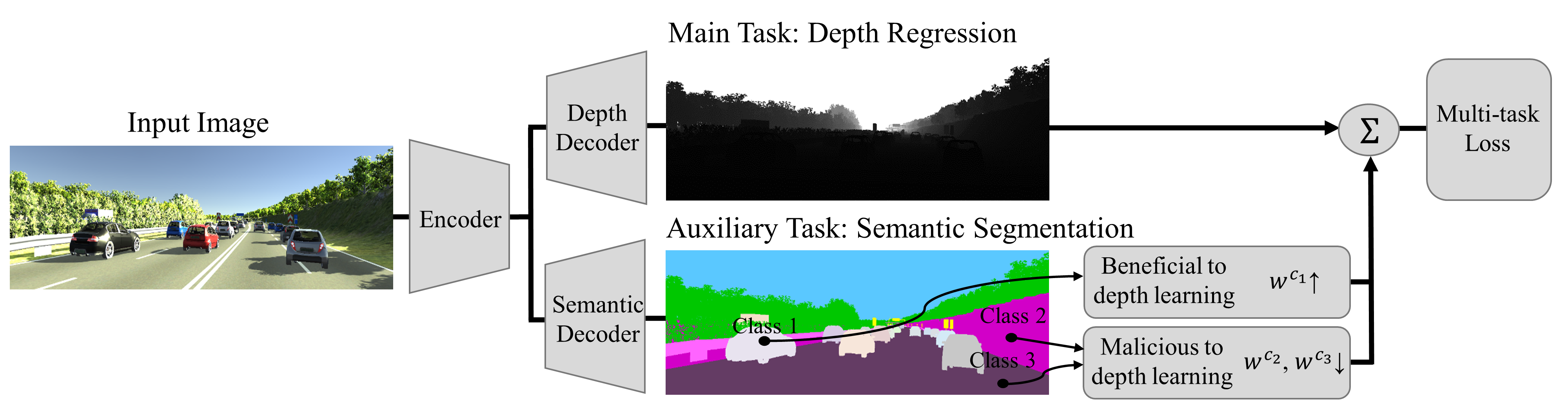}
  \caption{{\bf Example of the idea: the effect of semantic information on depth regression task.} labels, as a unit of information from the second task, are learned and affect the depth learning. In the process, both tasks conflict over some labels; for example, class "2", $c_2$, and class "3", $c_3$. At the moment, by decreasing the effect of $c_2$ and $c_3$ by reducing the multiplying-weight $w^{c_2}$ and $w^{c_3}$, the conflict can be minimized.}
\label{fig:fig1}
\end{figure}

\begin{table}[ht]
  \caption{{\bf The consequences of applying MTL depending on semantic classes.} An experiment to observe how MTL affects the accuracy by labels. Semantic segmentation and depth regression tasks are performed. Losses from two tasks are simply summed. Then, the depth regression accuracy is observed to see how it changes by segmentation labels. By the unit of information, some information (in this case, "car") is helpful, while some others are not.}
  \label{table:task_conflict_analysis}
  \centering
  \begin{tabular}{cllllll}
    \toprule
    & \multicolumn{2}{c}{Full Image} & \multicolumn{2}{c}{Non-car region} & \multicolumn{2}{c}{Car region}             \\
    \cmidrule(r){2-3} \cmidrule(r){4-5} \cmidrule(r){6-7}
    Metric & Single task & Multi task & Single task & Multi task & Single task & Multi task\\
    \midrule
    Acc. IoU(\%) & {\bf 98.29 } & 98.08 & {\bf 98.57 } & 98.32 & 87.03 & {\bf 88.18 }\\
    Log L1 & {\bf 0.0115 } & 0.0124 & {\bf 0.0104 } & 0.0115 & 0.0314 & {\bf 0.0283 }\\
    \bottomrule
  \end{tabular}
\end{table}

With the MTL setting, the model shares the prior information from the shared part of the model. Since the shared part learns from two tasks in the two-task-learning environment, the total information of the shared part increases. However, such information gain is not always positive for each other. While some information learned from some class labels of the second task has a positive effect on the first task, on the other hand, some others have a negative effect. For example, in Fig.~\ref{fig:fig1}, learning the labels of class "1" might be beneficial to depth learning, but learning the label "2" and "3" might be malicious to depth learning.
To verify the effect of task information, we performed a simple experiment with the two-task setting; semantic segmentation and depth regression. Then we analyzed the final accuracy of depth regression before and after MTL setting by semantic class labels. 

As shown in Table~\ref{table:task_conflict_analysis}, in the experiment, the accuracy of depth regression in the class label "car" region has increased, whereas some other class labels showed degraded depth accuracy. The result indicates that some class labels from semantic segmentation tasks harmed the depth regression task. To deal with such negative transfer, previous works suggested a weighted sum of losses, 
\begin{align}
   Loss = w_{t1}L_{t1} + w_{t2}L_{t2}
\label{eq:weighted_sum_loss}
\end{align}
and balanced the weights to find Pareto optimality between tasks. However, while this solution grants the presence of negative transfer, the suggested weighted sum does not remove negative transfer but is an ad-hoc solution to reduce it. 

Based on the analysis, this paper suggests class-wise weights on the auxiliary second task to draw a positive transfer to the main task. In the following chapters, this paper shows that the class-wise weights produce a positive transfer of information by updating the weights positively and reduce negative transfer otherwise.

\section{Iterative Reweighted Attention for Exploiting the Auxiliary Task}

Some information from the second task has a positive transfer, and others have a negative transfer to the first task. To find the optimality of the first task with the exploitation of information gain, this paper sets the second task as an auxiliary task and fully utilizes the information to boost the performance of the first (main) task. Then, to determine the positive and negative information from the auxiliary to the main task, we adopt class-wise weights ($w^c$) for each class label $c$, which are multiplied to the objective function of the auxiliary task. Therefore, the equation becomes 
\begin{align}
   Loss = L_{aux} + L_{main}
\label{eq:simple_sum_loss}
\end{align}
and
\begin{align}
   L_{aux} &= \sum_{c \in C}L_{aux}^c \\
   &= \sum_{c \in C}(w^c \times \frac{1}{N} \sum_{\forall i, y_i=c}D_{aux}(y_i, o_i)) \text{,}
\label{eq:loss_of_aux_task}
\end{align}
where $C =\{c_1,c_2,…,c_K\}$ for $K$ number of classes, $N$ is the number of $y$, and $D_{aux}$ is a distance function that calculates the distance between the true label, $y_i$, and the function output, $o_i$, at the $i$th point. The number of $i$ corresponds to the dimension of the output.

To update the class-wise weights, this paper suggests online iterative updates while training the multi-task function. To this end, one of the common updating methods is the gradient descent algorithm. Then the formula becomes 
\begin{align}
   w^c=w^c- \alpha \frac{\partial L_{main}^c}{\partial w^c } \text{,}
\label{eq:newton_step}
\end{align}
where $\alpha$ is a step size, and $L_{main}^c$ is the loss calculated over the points where its label of the second task is $c$. This updating rule indicates that $w^c$ is being updated to enhance the main task performance. With the optimal value of $w^c$, the main task fully exploits the information from the auxiliary task, drawing positive transfers only. However, since $L_{main}^c$ is not directly related to $w^c$, calculating $\sfrac{\partial L_{main}^c}{\partial w^c}$ is not possible yet. Instead, we reformulate it using chain rule: \begin{align}
   \frac{\partial L_{main}^c}{\partial w^c} = \frac{\partial L_{main}^c}{\partial L_{aux}^c} \times \frac{\partial L_{aux}^c}{\partial w^c} \text{.}
\label{eq:chain_rule}
\end{align}
Then, $\sfrac{\partial L_{aux}^c}{\partial w^c}$ can be directly calculated from Eq.~\eqref{eq:loss_of_aux_task}. However, $\sfrac{\partial L_{main}^c}{\partial L_{aux}^c}$ still cannot be resolved. Alternatively, we suggest an approximation by adopting a time sequence by the batch of inputs. Iteratively, at time $t$, a new batch of inputs is used to obtain $L(t)$, and the previous $L(t)$ becomes $L(t-1)$. Therefore, the equation becomes 
\begin{align}
   \frac{\partial L_{main}^c}{\partial L_{aux}^c} \approx  \frac{L_{main}^c(t)-L_{main}^c(t-1)}{L_{aux}^c(t)-L_{aux}^c(t-1)} \text{.}
\label{eq:approximation_using_time_sequence_of_loss}
\end{align}
Hence, the mathematical formula to update the weight becomes 
\begin{align}
   \frac{\partial L_{main}^c}{\partial w^c } \approx  \frac{L_{main}^c(t)-L_{main}^c(t-1)}{L_{aux}^c(t)-L_{aux}^c(t-1)} \times
   \frac{1}{N} \sum_{\forall i, y_i=c}D_{aux}(y_i, o_i) \text{.}
\label{eq:naive_solution_ours}
\end{align}
Although this solution might not have mathematical errors, there are some techniques to stabilize training steps in practice. For better convergence of weights, weights should not be changing swiftly. Hence, to grant weight momentum and stability, we refine the algorithm by adopting Adam optimizer~\cite{kingma2014adam} to the weight-updating gradient. Also, we added a small constant $\epsilon$ for the case that the equation would have a divide-by-zero problem. With a few more techniques, the final equation to update class-wise weights becomes 
\begin{align}
   w^c = \text{ReLU}\left(w^c - \alpha \frac{\widehat{m_t}}{\sqrt{\widehat{v_t}}+\epsilon}\right)
\label{eq:ours_adam_weight_update}
\end{align}
\begin{align}
   g_t = \frac{\partial L_{main}^c}{\partial w^c } \approx  \frac{\frac{L_{main}^c(t)-L_{main}^c(t-1)}{L_{main}^c(0)} + \epsilon}{\frac{L_{aux}^c(t)-L_{aux}^c(t-1)}{L_{aux}^c(0)} + \epsilon} \times \frac{1}{N} \sum_{\forall i, y_i=c}D_{aux}(y_i, o_i) \text{,}
\label{eq:final_solution_ours}
\end{align}
where $\widehat{m_t}$, $\widehat{v_t}$, and $g_t$ are the same variables in the Adam optimization algorithm~\cite{kingma2014adam}. With this equation, weights are stably being updated during the training of the model. The unit of time $t$ can be either a batch or a set of batches. We verified that the training process is stabilized with a batch size of 16. The overall algorithm of the proposed method is summarized in Algorithm~\ref{alg:algorithm1}.

During the training process, the algorithm boosts the main task's performance by updating the class-wise weights automatically in the auxiliary task. The updated weights affect the learning process through Eq.~\eqref{eq:loss_of_aux_task} and Eq.~\eqref{eq:simple_sum_loss}. To be more specific, in the case that some class labels would harm the main task learning, or that the main task has already exploited all useful information from them, their weights would converge to near zero, thereby preventing them from 
disrupting the main task learning. On the other hand, if some labels from the auxiliary task would be helpful for the main task learning, their weights would indicate the degree of importance for the main task learning, thereby affecting the learning process through the back-propagation of losses.

\begin{algorithm}
	\caption{Update equations for the proposed method} 
	\begin{algorithmic}[1]
	    \State Prepare the multi-task function $f_\theta$ optionally with pre-trained model parameter $\theta$
	    \State Set auxiliary and main task
	    \State Initialize class-wise weight $w^c = (w^{c_1},\ldots,w^{c_K}) \leftarrow (1,\ldots,1)$
	    \State Initialize step size $\alpha$, time step $t \leftarrow 0$
	    \While {$f_\theta$ not converged}
	        \State Get the function output $o$
	        \State Calculate $L = L_{aux} + L_{main}$ using Eq.~\eqref{eq:loss_of_aux_task}
	        \State Optimize $f_\theta$ in terms of $\theta$
	        
	        \State Calculate $g_t$ using Eq.~\eqref{eq:final_solution_ours}
	        \State Update $w_c$ using Eq.~\eqref{eq:ours_adam_weight_update}
	    
	        \State $t \leftarrow t+1$\
	    \EndWhile
	    \State Change the role of auxiliary and main task and repeat $line 2 - line12$
	\end{algorithmic} 
    \label{alg:algorithm1}
\end{algorithm}

\section{Experiments}

\subsection{Comparison with Previous Work}

In this chapter, we compare the proposed method with single-task models and the most promising previous work, Pareto MTL~\cite{lin2019pareto}, which is the generalized algorithm of its prior method, MOO-MTL~\cite{sener2018multi}. Pareto MTL~\cite{lin2019pareto} not only presents the best results but also supports an option to choose the task that we want to optimize. Hence, it is the most similar method to ours, among many other methods in the MTL field.

For pair comparison, we first prepared the U-Net~\cite{ronneberger2015u} as a base function, duplicated decoder branch for both of ours and Pareto MTL, and trained it with equal learning rate until the loss converge. For the MTL scenario, we prepared Virtual KITTI (VKITTI)~\cite{Gaidon:Virtual:CVPR2016}, NYU-v2~\cite{Silberman:ECCV12}, and Cityscapes~\cite{cordts2016cityscapes} datasets, all of which include semantic segmentation and depth regression labels. Then, we split VKITTI dataset to 9:1 for the training and testing purpose. NYU-v2 dataset has 795 training and 654 validation images. Cityscapes dataset has 2975 training and 500 validation images. We first set depth regression as the main task and semantic segmentation as an auxiliary task to obtain the best depth accuracy by updating class-wise weight in the semantic domain. 
Reversely, when we targeted semantic segmentation as the main task and placed depth regression as an auxiliary task, we quantized depth labels and assigned weights to get the best segmentation performance.

\begin{figure}[ht!]
\centering
\begin{minipage}[c]{.41\linewidth}
   \vspace{0.15cm}
   \strut\vspace*{-\baselineskip}\newline\includegraphics[width=1.05\linewidth]{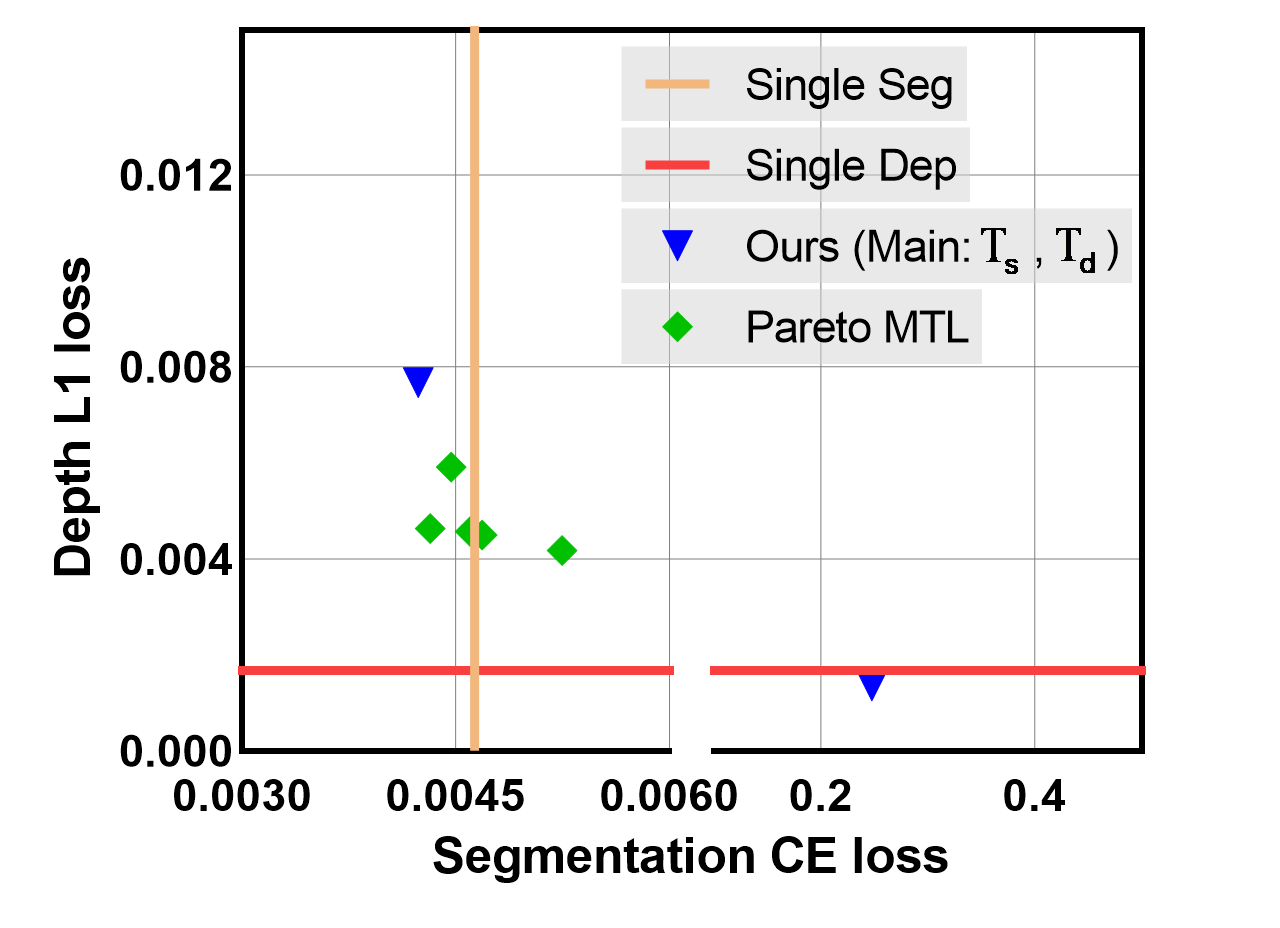} \label{fig:VKITTI}
\end{minipage}
\begin{minipage}[c]{.5\linewidth}
  \begin{tabular}{ccc}
    \toprule
    & $T_s$ Loss($10^{-2}$) & $T_d$ Loss($10^{-2}$) \\
    \midrule
    Single $T_s$ & 0.4632 &            \\
    Single $T_d$ &          &0.1682     \\
    \midrule
    Ours(Main: $T_s$) & \bf{0.4238} & 0.7672\\
    Ours(Main: $T_d$) & 24.7507 & \bf{0.1356}\\
    \midrule
    Pareto (0.0,1.0) &  0.4322 & 0.4631\\
    Pareto (0.92,0.38) & 0.4686 & 0.4498\\
    Pareto (0.71,0.71) & 0.4601 & 0.4575\\
    Pareto (0.38,0.92) & 0.5246 & 0.4173\\
    Pareto (1.0,0.0) & 0.4468 & 0.5918\\
    \bottomrule
  \end{tabular} \label{table:VKITTI}
\end{minipage}\hspace{1cm}
  \caption{{\bf Results of the experiment on the Virtual KITTI dataset.} $T_s$ and $T_d$ denote semantic segmentation and depth regression tasks each. Cross-entropy (CE) and L1 losses are used to measure $T_s$ and $T_d$ each on the validation set. The numbers in the bracket of Pareto MTL~\cite{lin2019pareto} represent reference vectors. Our proposed method shows the best performance for the target task. Pareto MTL suggests a set of Pareto optimal points with quite low depth and segmentation losses. However, it fails to enhance the depth regression task with the hindrance of the segmentation task.} 
  \label{fig:fig2}
\end{figure}

\begin{figure}[ht!]
\centering
\begin{minipage}[c]{.4\linewidth}
   \vspace{0.2cm} 
   \strut\vspace*{-\baselineskip}\newline\includegraphics[width=1.1\linewidth]{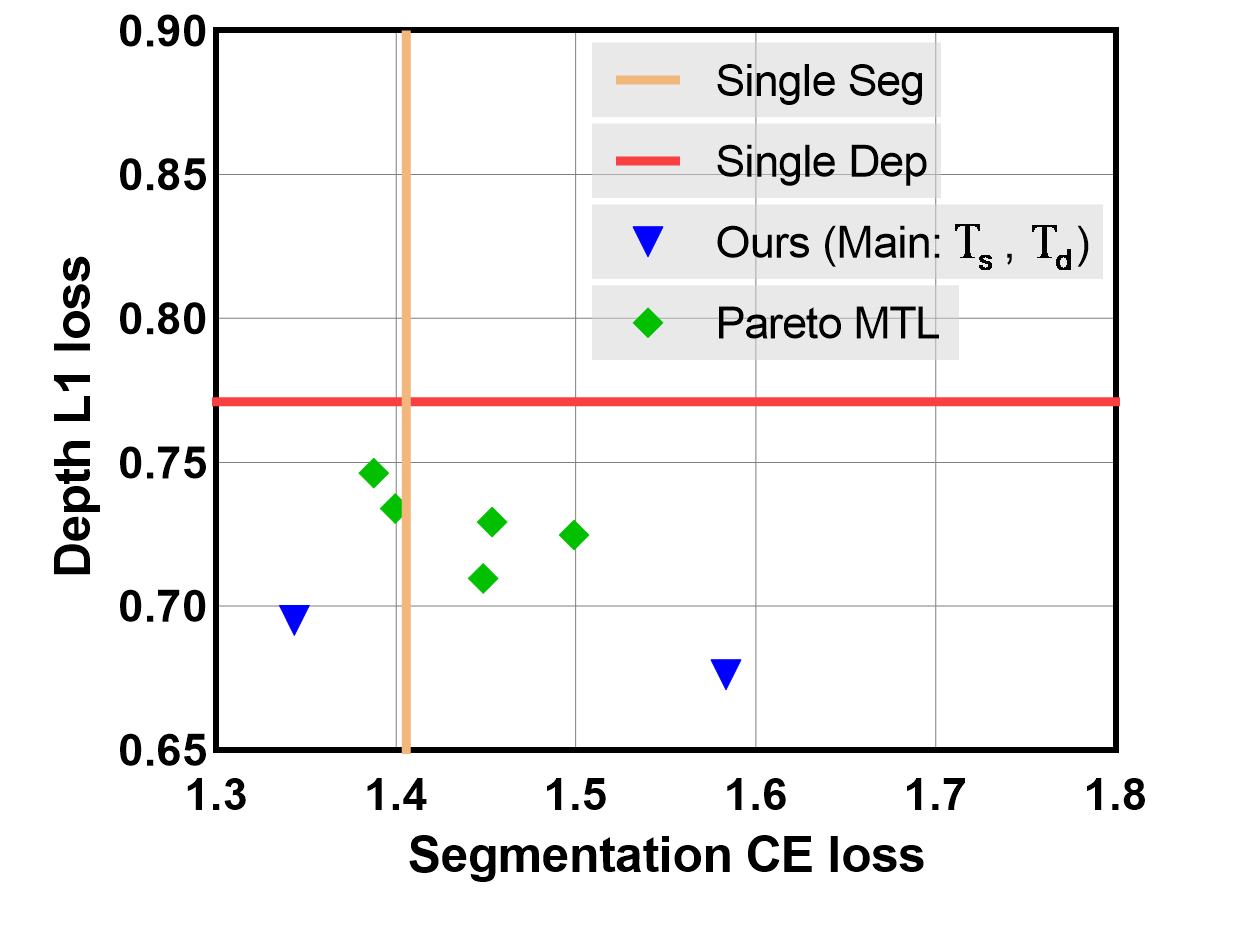} \label{fig:NYUV2}
\end{minipage}
\hspace{0.06\linewidth}
\begin{minipage}[c]{.4\linewidth}
  \begin{tabular}{ccc}
    \toprule
    & $T_s$ Loss & $T_d$ Loss \\
    \midrule
    Single $T_s$ & 1.4056 &            \\
    Single $T_d$ &          & 0.7711     \\
    \midrule
    Ours(Main: $T_s$) & \bf{1.3436} & 0.6951\\
    Ours(Main: $T_d$) & 1.5835 & \bf{0.6761}\\
    \midrule
    Pareto (0.0,1.0) & 1.3876 & 0.7462\\
    Pareto (0.92,0.38) & 1.3995 & 0.7340\\
    Pareto (0.71,0.71) & 1.4535 & 0.7293\\
    Pareto (0.38,0.92) & 1.4485 & 0.7096\\
    Pareto (1.0,0.0) & 1.4989 & 0.7248\\
    \bottomrule
  \end{tabular} \label{table:NYUV2}
\end{minipage}\hspace{1cm}
  \caption{{\bf Results of the experiment on the NYU-v2 dataset.} Our proposed model can find the optimal points with the lowest depth and segmentation losses.}
  \label{fig:fig3}
\end{figure}

\begin{figure}[ht!]
\centering
\begin{minipage}[c]{.4\linewidth}
   \vspace{0.13cm}
   \strut\vspace*{-\baselineskip}\newline\includegraphics[width=1.1\linewidth]{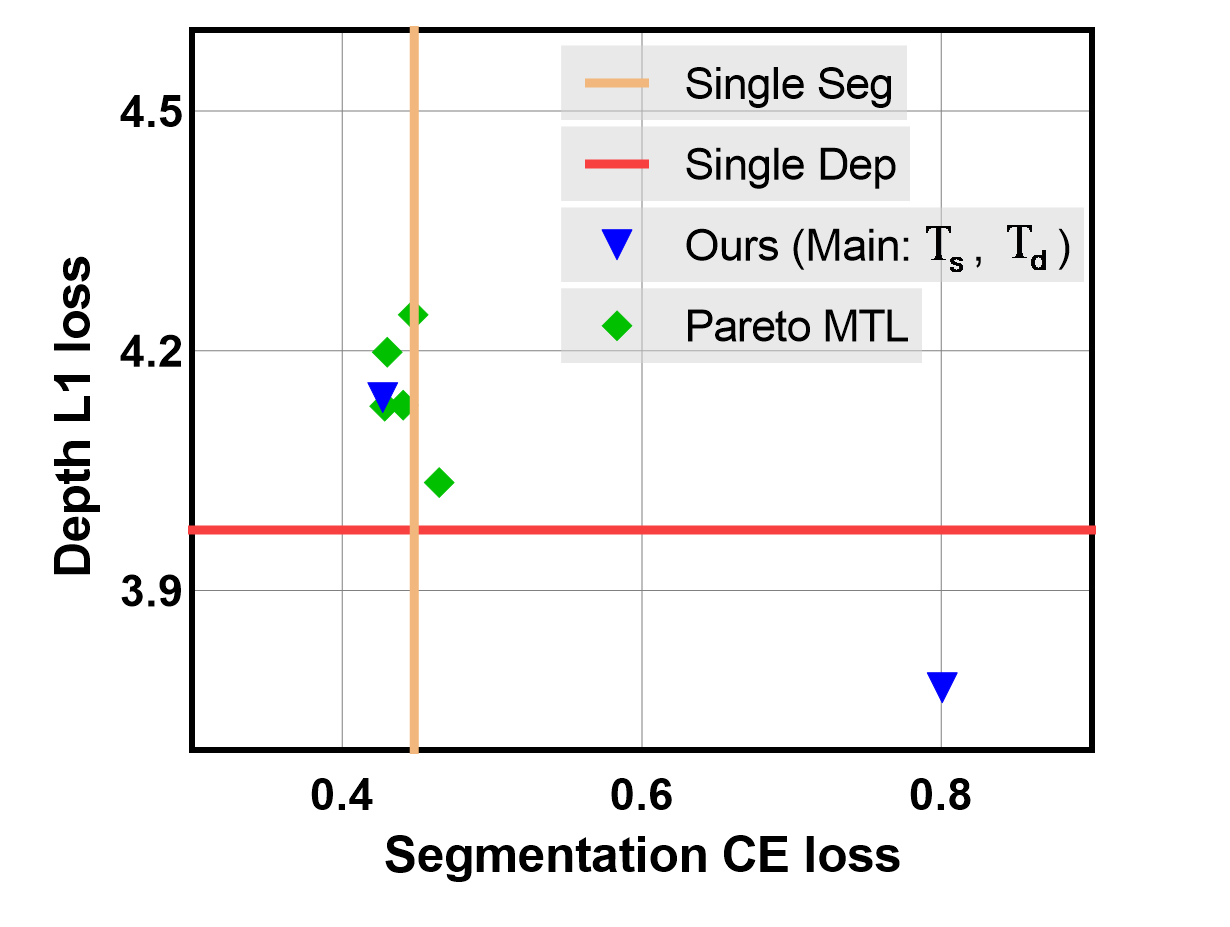} \label{fig:Cityscapes}
\end{minipage}
\hspace{0.06\linewidth}
\begin{minipage}[c]{.4\linewidth}
  \begin{tabular}{ccc}
    \toprule
    & $T_s$ Loss & $T_d$ Loss \\
    \midrule
    Single $T_s$ & 0.4482 &            \\
    Single $T_d$ &          &3.9754     \\
    \midrule
    Ours(Main: $T_s$) & \bf{0.4270} & 4.1408 \\
    Ours(Main: $T_d$) & 0.8007 & \bf{3.7781}\\
    \midrule
    Pareto (0.0,1.0) & 0.4302 & 4.1978 \\
    Pareto (0.92,0.38) & 0.4648 & 4.0350\\
    Pareto (0.71,0.71) & 0.4408 & 4.1314\\
    Pareto (0.38,0.92) & 0.4284 &4.1305\\
    Pareto (1.0,0.0) & 0.4475 &  4.2447\\
    \bottomrule
  \end{tabular}  \label{table:Cityscapes}
\end{minipage}\hspace{1cm}
  \caption{{\bf Results of the experiment on the Cityscapes dataset.} The optimal point from our proposed model shows the biggest gap in depth-regression loss among all other experiments. Pareto MTL tends to enhance semantic segmentation, but not the depth regression. From the observation in this dataset, the semantic segmentation task can generally gain positive information from the depth regression task, but not otherwise. Therefore, depth regression can be enhanced only by selectively exploiting the positive information from the semantic segmentation label.}
  \label{fig:fig4}
\end{figure}

In all experiments, the best segmentation and the best depth losses estimated from the proposed method consistently achieved the lowest validation losses. Although Pareto MTL~\cite{lin2019pareto} is known to suggest a set of Pareto optimal points with an appropriate trade-off between different tasks, the sum of weighted losses reveals weak performance on a few datasets where the task conflicts are dominant referring to Fig.~\ref{fig:fig2} and Fig.~\ref{fig:fig4}.

As mentioned above, we proposed a new solution to boost the performance of the one target task on which we focus. That is, our model usually costs the auxiliary tasks for the main one. Because of this one-side learning tendency, optimal points obtained from our model shows high depth loss when we focus on segmentation task and marks high segmentation loss when we concentrate on depth task. Even though optimal points of ours in Fig.~\ref{fig:fig3} exceptionally achieved excellent performance in both tasks at the same time, our optimalities commonly have highly outstanding performance in one main task by sacrificing the performances in other tasks. Compared to previous MTL papers that tried to get high accuracy in most tasks simultaneously, we concentrate on boosting the learning of the one main task from the viewpoint of MTL. Thereby, our model could be regarded as a 'real-time' transfer learning as it automatically invigorates positive transfer and suppresses negative transfer between tasks during the training of the model.

Moreover, during the training, it does not require complex computation for updating the weights. As shown in Table.~\ref{table:runtime}, our algorithm runs almost as fast as a simple linear sum of losses. These properties make the proposed method easily extendable to other applications in practice, such as a mobile phone cameras or autonomous driving.

\begin{table}
  \caption{{\bf Run-time comparison.} We measured the average time to process one batch of 8 images. Tests were done using the machine with Python Numpy, Nvidia GTX 1080ti, and Intel i7-8700 CPU.}
  \label{table:runtime}
  \centering
  \begin{tabular}{llll}
    \toprule
    & Linear Sum & Pareto MTL & \bf{Ours} \\
    \midrule
    Run-time (second) & 0.183 & 1.128 & \bf{0.310} \\
    \bottomrule
  \end{tabular}
\end{table}

\subsection{Details about Experiments}

For pair comparison, we initially prepared an equal environment for all methods. Firstly, we used U-Net~\cite{ronneberger2015u} as a baseline in all models. Referring to previous MTL papers in visual tasks, we used cross-entropy loss for semantic segmentation and L1 loss for depth regression. We also preprocessed depth labels to inverse depth labels for VKITTI dataset. During the training of each method, we fixed the learning rate as 0.01 and then decreased it as 0.001 after 20 epochs. Then, we collected the best results from each method. 

In the proposed method, all class-wise weights are initialized as one and updated after 20 (for VKITTI and NYU-v2) and 12 (for Cityscapes) epochs using Eq.~\eqref{eq:newton_step} and Eq.~\eqref{eq:final_solution_ours}. Note that we deliberately fixed class-wise weights for stability at the initial stage. Then, depending on the size of the dataset, we chose different starting points to update weight. In the same way, step size $\alpha$ in Eq.~\eqref{eq:newton_step} depends on the size of dataset. For easy tasks where the dataset is small, and the model might overfit fast, such as Cityscapes dataset, a large value of $\alpha$ would be suitable, while a small value of 0.0002 generally works well.

\section{Conclusion and Future Work}

In this paper, a new method is presented in MTL problems to boost the performance of the main task with the exploitation of the auxiliary task. The proposed method suggests class-wise weights to balance the internal effect within a task to help main task learning. In the experiment, the proposed method records the best performance for the target task and runs almost as fast as a simple linear sum of losses. Once the model is trained, the auxiliary task can be omitted. The trained weights can be used as a single-task model in its application, but with improved accuracy. Although we tested our approach with two popular vision tasks, the proposed method can be generally applied to areas wherever the dataset has multi-labels. The continuous domain of labels can be discretized to assign class-wise labels. Locations, signals, and probability can be discretized as well. 

Lastly, we are planning to automate initial weights and other adjustable hyperparameters so that the model can be applied to other problems immediately. It is also planned to generalize our experiment in the future, combining with recent knowledge in distillation learning techniques.

\section*{Broader Impact}

This work has the following potential positive impact in the society; a) Autonomous driving perception model may have benefit in its performance using multiple cameras.

This work does not present any foreseeable negative societal consequence.

This work does not leverage any bias in the data.

\begin{ack}
We thank our teammates for supporting this work.
\end{ack}

\small

\bibliographystyle{acm}
\bibliography{MTL_refs}

\end{document}